\documentclass{article}

\usepackage{arxiv}

\usepackage[utf8]{inputenc} % allow utf-8 input
\usepackage[T1]{fontenc}    % use 8-bit T1 fonts
\usepackage{hyperref}       % hyperlinks
\usepackage{url}            % simple URL typesetting
\usepackage{booktabs}       % professional-quality tables
\usepackage{amsfonts}       % blackboard math symbols
\usepackage{nicefrac}       % compact symbols for 1/2, etc.
\usepackage{microtype}      % microtypography
\usepackage{lipsum}		% Can be removed after putting your text content
\usepackage{graphicx}
\usepackage{doi}
\usepackage{amssymb,amsmath,nccmath}  
\usepackage{macros}
\usepackage{xcolor}
\usepackage{multirow}
\usepackage[normalem]{ulem}

\title{Consistency-based Semi-supervised Evidential Active Learning for Diagnostic Radiograph Classification}

\author{Shafa Balaram$^{1,2}$
\quad
Cuong M. Nguyen$^2$
\quad 
Ashraf Kassim$^3$\thanks{Equal Contribution}
\quad Pavitra Krishnaswamy$^{2*}$ \\ \\
$^1$National University of Singapore, Singapore \\
\texttt{shafa.balaram@u.nus.edu} \\
$^2$Institute for Infocomm Research, A*STAR, Singapore \\
\texttt{pavitrak@i2r.a-star.edu.sg} \\
$^3$Singapore University of Technology and Design, Singapore
}

\renewcommand{\shorttitle}{Consistency-based Semi-supervised Evidential Active Learning}

\hypersetup{
pdftitle={Consistency-based Semi-supervised Evidential Active Learning for Diagnostic Radiograph Classification},
pdfauthor={Shafa Balaram, Cuong M. Nguyen, Ashraf Kassim, Pavitra Krishnaswamy},
pdfkeywords={Semi-supervised learning, Active learning, Theory of evidence, Subjective logic, Multi-label classification.},
}

\begin{document}
\maketitle

\begin{abstract}
Deep learning approaches achieve state-of-the-art performance for classifying radiology images, but rely on large labelled datasets that require resource-intensive annotation by specialists. 
Both semi-supervised learning and active learning can be utilised to mitigate this annotation burden. However, there is limited work on combining the advantages of semi-supervised and active learning approaches for multi-label medical image classification. Here, we introduce a novel Consistency-based Semi-supervised Evidential Active Learning framework (CSEAL). Specifically, we leverage predictive uncertainty based on theories of evidence and subjective logic to develop an end-to-end integrated approach that combines consistency-based semi-supervised learning with uncertainty-based active learning. We apply our approach to enhance four leading consistency-based semi-supervised learning methods: \textit{Pseudo-labelling}, \textit{Virtual Adversarial Training}, \textit{Mean Teacher} and \textit{NoTeacher}. Extensive evaluations on multi-label Chest X-Ray classification tasks demonstrate that CSEAL achieves substantive performance improvements over two leading semi-supervised active learning baselines. Further, a class-wise breakdown of results shows that our approach can substantially improve accuracy on rarer abnormalities with fewer labelled samples.\let\thefootnote\relax\footnotetext{Preprint submitted to MICCAI. Accepted in May 2022.}
\end{abstract}

% keywords can be removed

\section{Introduction}
Deep learning approaches offer leading-edge performance for automated image classification applications in the domain of radiology~\cite{rajpurkar2018deep,flanders2020construction,mckinney2020international}. However, training deep learning models requires large datasets that are carefully labelled by clinical specialists. In practice, expert annotation of large-scale medical image databases is highly laborious, resource-intensive, and often infeasible.  

To address the annotation challenge, it is desirable to (a) prioritise labelling of the most informative images and (b) simultaneously leverage large unlabelled image collections that are readily available within routine clinical databases for model training. However, each of these objectives requires distinct approaches: (a) active learning to select images for labelling, and (b) semi-supervised learning to leverage limited labelled datasets alongside larger unlabelled datasets. Semi-supervised active learning strategies combine advantages of the two approaches to generate better feature representations, even when starting with small randomly sampled labelled sets~\cite{gao2020consistency,borsos2021semi}. Although these approaches have been explored for many applications involving natural scene images, there are limited demonstrations of semi-supervised active learning for diverse medical image classification tasks~\cite{boushehri2021systematic,yang2020deep,zotova2019comparison}.

Here, we propose a novel Consistency-based Semi-supervised Evidential Active Learning (CSEAL) framework that learns and leverages predictive uncertainty alongside the classification objective in an end-to-end manner. Our approach builds upon the Dempster-Shafer theory of evidence~\cite{dempster1968generalization} and the principles of subjective logic~\cite{jsang2016subjective}, concurrently estimates the label uncertainty and prediction consistency to facilitate active learning and semi-supervised learning respectively, and is well-suited to handle the diverse challenges for radiology image classification. Our main contributions are as follows:
\begin{enumerate}
\item We propose a new end-to-end integrated approach to combine consistency-based semi-supervised learning with uncertainty-based active learning, by learning and utilising the class-wise evidences for both parameter optimisation and uncertainty estimation. 
\item We apply CSEAL to develop evidential analogues of four leading consistency-based semi-supervised methods: Pseudo-labelling (PSU)~\cite{lee2013pseudo}, Virtual Adversarial Training (VAT)~\cite{miyato2018virtual}, Mean Teacher (MT)~\cite{tarvainen2017mean}, and NoTeacher (NoT)~\cite{unnikrishnan2021semi}. 
\item In realistic experiments on the NIH-14 Chest X-Ray dataset~\cite{wang2017hospital}, we demonstrate that CSEAL outperforms leading semi-supervised active learning baselines with very low labelling budgets of under 5\%. In particular, for rarer abnormalities with $<5\%$ prevalence, we observe that CSEAL enables substantial performance gains of up to 17\% in AUROC over evidential supervised learning with random sampling. 
\end{enumerate}

%%%%%%%%%%%%%%%%%%%%%%%%%%%%%%%%%%%%%%%%%%%%%%%%%%%%%%%%

\section{Related Works}

We briefly review two recent state-of-the-art semi-supervised active learning approaches characterised in this work. First,~\cite{huang2021semi} proposed an unlabelled sample loss estimation method which applies Temporal Output Discrepancy (TOD) for semi-supervised learning and Cyclic Output Discrepancy (COD) for active learning. This TOD+COD method compared the change in model predictions between optimisation steps based on MT and active learning cycles. Second, Virtual Adversarial Training with Augmentation Variance (VAT+AugVar)~\cite{gao2020consistency}, an augmentation-based active learning approach, quantifies informativeness using the variance across augmented sample predictions as demonstrated with VAT. 

Other salient semi-supervised active learning approaches in the computer vision literature include Cost-Effective Active Learning (CEAL), and MixMatch-based methods. CEAL incorporated PSU in multiple uncertainty-based active learning heuristics~\cite{wang2016cost}, but this approach can propagate label noise during model training and also exhibit uncertainty in network predictions even when the probability output after softmax is high~\cite{gal2016uncertainty}. MixMatch, a dominant semi-supervised learning method, was combined separately with various active learning techniques such as data summarisation~\cite{borsos2021semi}, label propagation~\cite{guo2021semi}, and k-means and cosine similarity distances~\cite{song2019combining}. However, the MixUp augmentation technique used in MixMatch limits its applicability to medical image classification tasks.

Some studies have also proposed semi-supervised active learning for medical image applications. For instance, CEAL~\cite{zotova2019comparison} was adapted with representativeness sampling for computerised tomography (CT) in lung nodule segmentation. Further, Co-training active learning (COAL)~\cite{yang2020deep} used a deep co-training with a hybrid active learning acquisition function for mammography image classification. Recently, PSU was combined with Batch Active learning by Diverse Gradient Embeddings (BADGE)~\cite{ash2019deep}, and demonstrated on three biomedical datasets~\cite{boushehri2021systematic}. Yet, there remains a need for a unified and customisable framework that can address diverse challenges in radiology image classification. For example, COAL's choice of active learning criteria is highly data and task specific, hindering its generalisability. Additionally, BADGE is not easily scalable to multi-label settings, due to its computationally expensive k-means clustering.

%%%%%%%%%%%%%%%%%%%%%%%%%%%%%%%%%%%%%%%%%%%%%%%%%%%%%%%%

\section{CSEAL Framework}
In this section, we provide an overview of our proposed CSEAL framework. We consider the semi-supervised active learning setup with held-out validation and test sets. The training labelled set and validation set are initialised by random sampling until the annotation budget is met or there is complete class coverage. The notations $\{\vxl_i, \vy_i\}_{i=1}^{L_T}$ and $\{\vxu_i\}_{i=1}^{L_U}$ are used to denote the labelled and remaining unlabelled training samples respectively. The validation set is denoted as $\{\vxl_i, \vy_i\}_{i=1}^{L_V}$. For a realistic process, we assume that $L_V \ll L_T$.

\begin{figure}[h!]
\includegraphics[width=\textwidth]{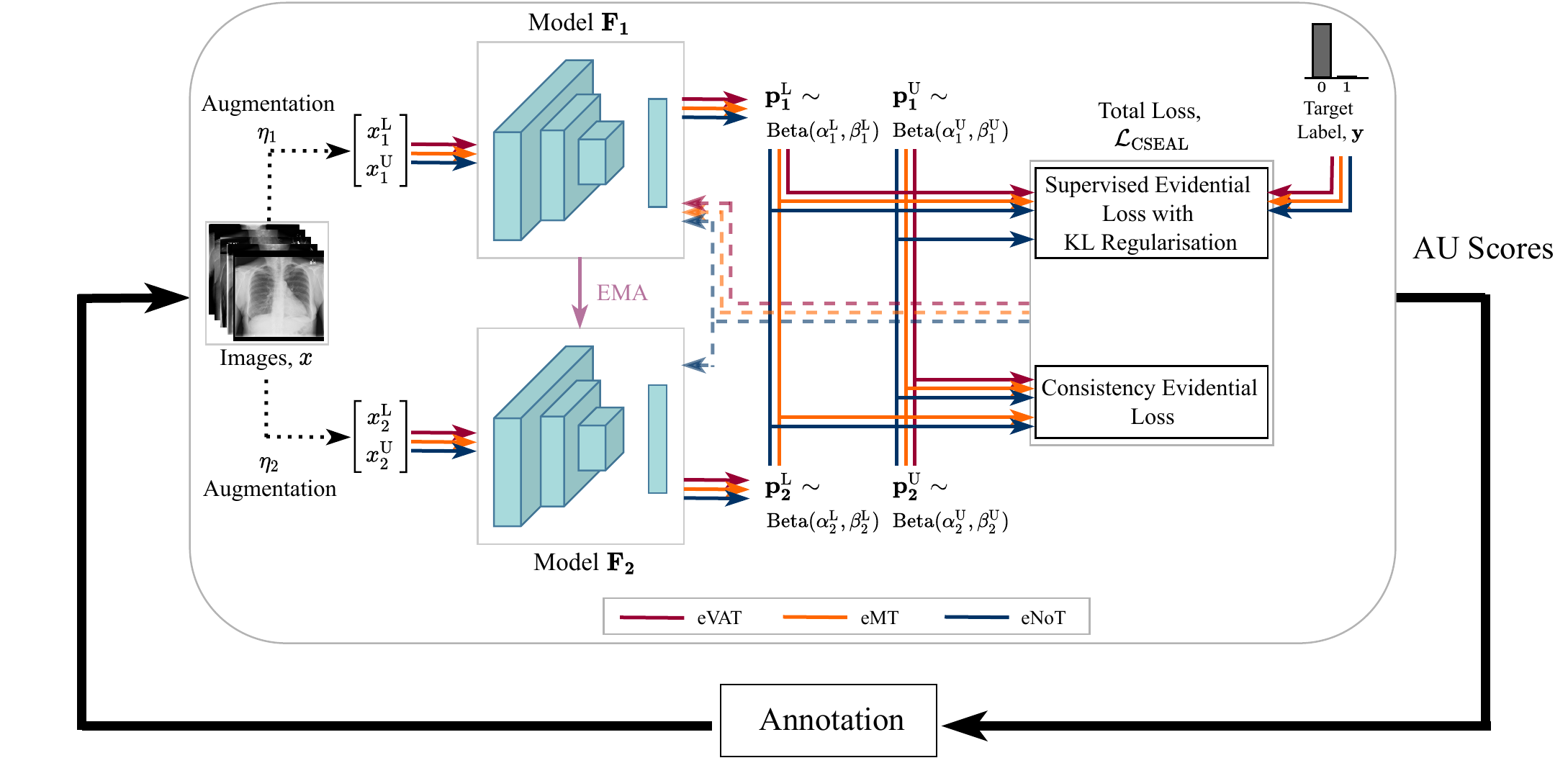} 
\caption{Overview of the Consistency-based Semi-supervised Evidential Active Learning (CSEAL) framework. Augmentations $\eta_1$ and $\eta_2$ are applied to the input image $x$ to generate $x_1$ and $x_2$. These are fed into the parametrised networks $F_1$ and $F_2$ before computing the supervised evidential loss and the consistency evidential loss. The forward and backward propagation are denoted by solid and dashed arrows respectively. The aleatoric uncertainty (AU) score is calculated for for each unlabelled image to prioritise them for annotation. CSEAL is shown here with the evidential analogues of the 3 most recent consistency-based semi-supervised learning methods, namely evidential Virtual Adversarial Training (eVAT), Mean Teacher (eMT) and NoTeacher (eNoT). In the case of ePSU, the pseudo-labels inferred by the network $F_1$ are used in the supervised evidential loss of the unlabelled samples.} 
\label{fig:cseal}
\end{figure}

Figure~\ref{fig:cseal} illustrates the Consistency-based Semi-supervised Active Learning (CSEAL) framework for binary classification. A multi-label classifier can be achieved by having multiple binary classification heads. The full version of CSEAL (with two networks) involves two major steps: (1) evidential-based semi-supervised learning and (2) evidential-based active learning. \\

\noindent \textbf{(1) Evidential-based Semi-supervised Learning:} Given a training image $x$, the transformation functions $\eta_1$ and $\eta_2$ are applied to the training images to generate the augmented samples $\{\vxl_1, \vxu_1\}$ and $\{\vxl_2, \vxu_2\}$ respectively. We apply two neural networks $F_1$ and $F_2$ to generate outputs from the corresponding augmented inputs. Since we are dealing with binary classification, the class predictors $\vp_1=[p^+_1, p^-_1]^\top$ and $\vp_2=[p^+_2, p^-_2]^\top$ can be obtained by applying a sigmoid function on the output logits $\bf{f}_1$ and $\bf{f}_2$. However, they are just point estimates which do not carry uncertainty information. Inspired from the evidential-based uncertainty estimation works~\cite{sensoy2018evidential,ghesu2021quantifying}, we assume that the Bernoulli variables $\vpone$ and $\vptwo$ have priors in the form of Beta distributions parameterised by $\valpha_1=[\alpha_1, \beta_1]$ and $\valpha_2=[\alpha_2, \beta_2]$, respectively. We use the output logits from the networks to compute evidences and estimate $\valpha_1$ and $\valpha_2$ using $\valpha=\mathrm{exp}(\bf{f})+\bf{1}$ with $\bf{f}$ clamped to $[-10, 10]$. Unlike a standard neural network classifier where the output is squashed into a probability assignment (or class predictor), i.e., $P(y=+) = p^{+}$, CSEAL uses network outputs to parameterise a Beta prior instead, which represents the density of each and every possible probability assignment. Hence, CSEAL models the second-order probabilities and uncertainty~\cite{jsang2016subjective}. The Beta prior also enables the estimation of aleatoric uncertainty, which will be discussed in the next step. 

At inference, the prediction probabilities of each class are computed as the mean of the Beta distribution, i.e. $\vhp_1 = [\hppos_1, \hpneg_1]^\top = [\alpha_1/E_1, \beta_1/E_1]^\top$ where the total evidence is computed as $E=\alpha+\beta$. Here, the term \textit{evidence} refers to the amount of support obtained from a data sample in favor of positive or negative predictions. Subsequently, we define a semi-supervised loss as a function of the class predictors and compute its Bayes risk w.r.t. Beta distribution priors. Additionally, a Kullback-Leibler (KL) divergence term is also included to penalise predictions with high uncertainty. Specifically, this KL term computes divergence between the Beta prior with adjusted parameters $\pmb{\tilde{\tau}}=\vy+(1-\vy)\odot \pmb{\tau}$ and the uniform Beta distribution which represents the state of complete uncertainty. The general loss of CSEAL is: 

\begin{align}
\label{eq:cseal_loss}
\begin{split}
\mathfrak{L}_{\CSEAL}(\vx, \vy) &= \lambda_{sup} \left[ \mathfrak{L}_{err}(\vy, \hat{\mathbf{p}}) + \mathfrak{L}_{var}(\hat{\mathbf{p}}, \pmb{\tau}) + \lambda_t\ \mathfrak{L}_{reg}(\valpha, \vy) \right] \\
&+ \lambda_{cons}  \mathfrak{L}_{cons}(\hat{\mathbf{p}_1}, \hat{\mathbf{p}_2})
\end{split}
\end{align}
where $\lambda_t=\min (1.0, t/10)$ is the adaptive regularisation coefficient over the first $t$ epochs. The supervised loss terms $\mathfrak{L}_{err}(\vy, \hat{\mathbf{p}})$ and $\mathfrak{L}_{var}(\hat{\mathbf{p}}, \pmb{\tau})$ originate from the Bayes risk of the squared error between $\vy$ and $\vp$. The term $\mathfrak{L}_{reg}(\valpha, \vy)$ is a result of regularisation using KL divergence. The (optional) consistency term $\mathfrak{L}_{cons}(\hat{\mathbf{p}_1}, \hat{\mathbf{p}_2})$ is only computed between the outputs of two separate networks.

When CSEAL is applied to SUP (a supervised learning baseline), PSU and VAT, we adopt a single-network architecture and drop the second network $F_2$. When applied to MT and NoT, CSEAL takes the full form with two separate networks. We now describe how $\mathfrak{L}_{\CSEAL}(\vx, \vy)$ can be adapted for different parametrised consistency-based semi-supervised learning models:
\begin{itemize}
    \item \textbf{ePSU:} We infer the pseudo-labels from the mean of the Beta distribution. 
    \item \textbf{eVAT:} We optimise the Bayes risk of the squared error between the class predictor on an unlabelled sample and its adversarial counterpart. 
    \item \textbf{eMT:} We take the Bayes risk of the squared error between the class predictors of the student and the EMA-updated teacher networks w.r.t. their respective Beta distribution priors. 
    \item \textbf{eNoT:} We optimise the Bayes risk of the squared error terms in the log likelihood w.r.t. their Beta distribution priors. 
\end{itemize}
We provide additional details about these loss functions in the Supplement. \\

\noindent \textbf{(2) Evidential-based Active Learning:} To facilitate active learning, we compute the aleatoric uncertainty (AU) as previous work has shown that AU is more effective than epistemic uncertainty (or model uncertainty) in evidential graphical semi-supervised learning~\cite{zhao2020uncertainty}.  We estimate AU for each class as the expected entropy of the class predictor $\vp$ given its Beta distribution prior as follows:
\begin{equation}
\label{eq:au_unc}
  \mathrm{AU} = \mathbb{E}_{p \sim Beta(\alpha, \beta)} \left \{ \mathcal{H} [p] \right \} = \frac{1}{\ln 2} \sum_{\gamma \in \{\alpha, \beta\}} \frac{\gamma}{E} \left(\psi(E+1) - \psi(\gamma+1) \right) 
\end{equation}
where $\psi(\cdot)$ is the \textit{digamma} function. The label-level AU scores are aggregated to obtain the image-level uncertainty score. 

We annotate the unlabelled images with the highest scores and use these to augment the labelled training set. We also randomly select and label additional validation samples such that the ratio $L_T:L_V$ is maintained throughout the active learning process. The networks $F_1$ and $F_2$ continue to be trained with backpropagation and the above steps are repeated as part of an iterative process until the final labelling budget is met.

%%%%%%%%%%%%%%%%%%%%%%%%%%%%%%%%%%%%%%%%%%%%%%%%%%%%%%%%

\section{Experiment Setup}
\noindent \textbf{Dataset}: We demonstrate our method on the NIH-14 Chest X-Ray dataset~\cite{wang2017hospital}. This dataset contains 112,120 high-dimensional frontal radiographs that are labelled for presence or absence of one or more abnormalities from 14 pathologies. The dataset exhibits high class imbalance. Out of the 46.1\% of images containing at least one abnormality, 40.1\% are multi-labelled. We use publicly available training (70\%), validation (10\%) and test (20\%) splits generated without patient overlaps~\cite{zech284reproduce}.

\noindent \textbf{Realistic Active Sampling Process and Labelling Regimes}: We ensured our active sampling process is reflective of practical clinical annotation workflows. First, as we used a separate validation pool, the labelled validation sets are representative of the held-out test set throughout the experiment, and sized to be much smaller than the training set. Second, our CSEAL-based annotation process is realistic as it does not require stratified or class-balanced initial labelled training sets or aligned training and validation class distributions. We performed experiments in two labelling regimes, namely: (a) the low-range regime from 2\% to 5\% in steps of 0.5\% and (b) the mid-range regime from 5\% to 10\% in steps of 1\%. 

\noindent \textbf{Experiments and Baselines}: We  evaluated the performance of ePSU, eVAT, eMT, and eNoT, and benchmarked against two competitive semi-supervised active learning methods, TOD+COD~\cite{huang2021semi} and VAT+AugVar~\cite{gao2020consistency}. We also included an evidential supervised baseline (eSUP) to assess gains arising from inclusion of the unlabelled data. All evidential methods are evaluated using both Random and AU sampling. We use the same held-out test set across all experiments for fair comparisons.

\noindent \textbf{Implementation Details}:
We describe the evidential supervised learning setup and hyperparameter tuning process. Additional details are in the Supplement. 

\paragraph{Model Training:} For fair comparisons across all the methods evaluated, we utilise the same DenseNet121 classifier backbone followed by a dropout layer and fully-connected layer as per~\cite{ghesu2021quantifying}. The logits of each class are mapped to the parameters $[\alpha, \beta]$ of the Beta distribution using an exponential activation function for the evidential classifiers, and to the sigmoidal prediction probabilities for the standard classifiers. After each annotation round, we reset the network to its pre-trained ImageNet weights for retraining. The input images are augmented using random affine transformations with rotation of $\pm 10^\circ$, translation of up to $10\%$ and scaling between 0.9 and 1.1 followed by random horizontal flipping, resizing and centre cropping during training. The transformed images are then normalised using ImageNet mean and standard deviation. We use an Adam optimiser ($\beta = [0.9, 0.999], \epsilon =1\times10^{-8}$) with learning rate of $1\times10^{-4}$, linear decay scheduler of $0.1$ based on the validation loss, and a weight decay of $1\times10^{-5}$. All methods are implemented in PyTorch v1.4.0~\cite{paszke2017automatic}. 

\noindent{\textbf{Hyperparameter Tuning:}} Since a complete grid search to find the optimal hyperparameters for each semi-supervised active learning method is computationally infeasible, we use optimal hyperparameter configurations employed in previous works on non-evidential semi-supervised learning~\cite{unnikrishnan2020semi,unnikrishnan2021semi} and tune the dropout rate for each method. For TOD+COD, we utilise the original implementation with EMA decay rate of 0.999 and consistency weight $\lambda=0.05$. 

\noindent{\textbf{EMA Averaging:}} We retain an exponential moving average (EMA) copy of the model weights for all methods to ensure a fairer comparison against eMT and TOD+COD. This enables us to attribute the performance improvements observed to parameter averaging, consistency mechanism and/or active sampling. The active learning scores are computed from the model or its EMA copy based on the validation AUROC. For all evaluations, we report the best test AUROC either from the model or from its EMA copy.

%%%%%%%%%%%%%%%%%%%%%%%%%%%%%%%%%%%%%%%%%%%%%%%%%%%%%%%%

\section{Results}
Here, we report performance of CSEAL in relation to the baselines in the low-range labelling regime. The Supplement includes results of CSEAL and the baselines in the mid-range labelling regime, as well as the performance comparison of the evidential semi-supervised learning approaches against their non-evidential counterparts.

\begin{figure}[h!]
\includegraphics[width=0.49\textwidth]{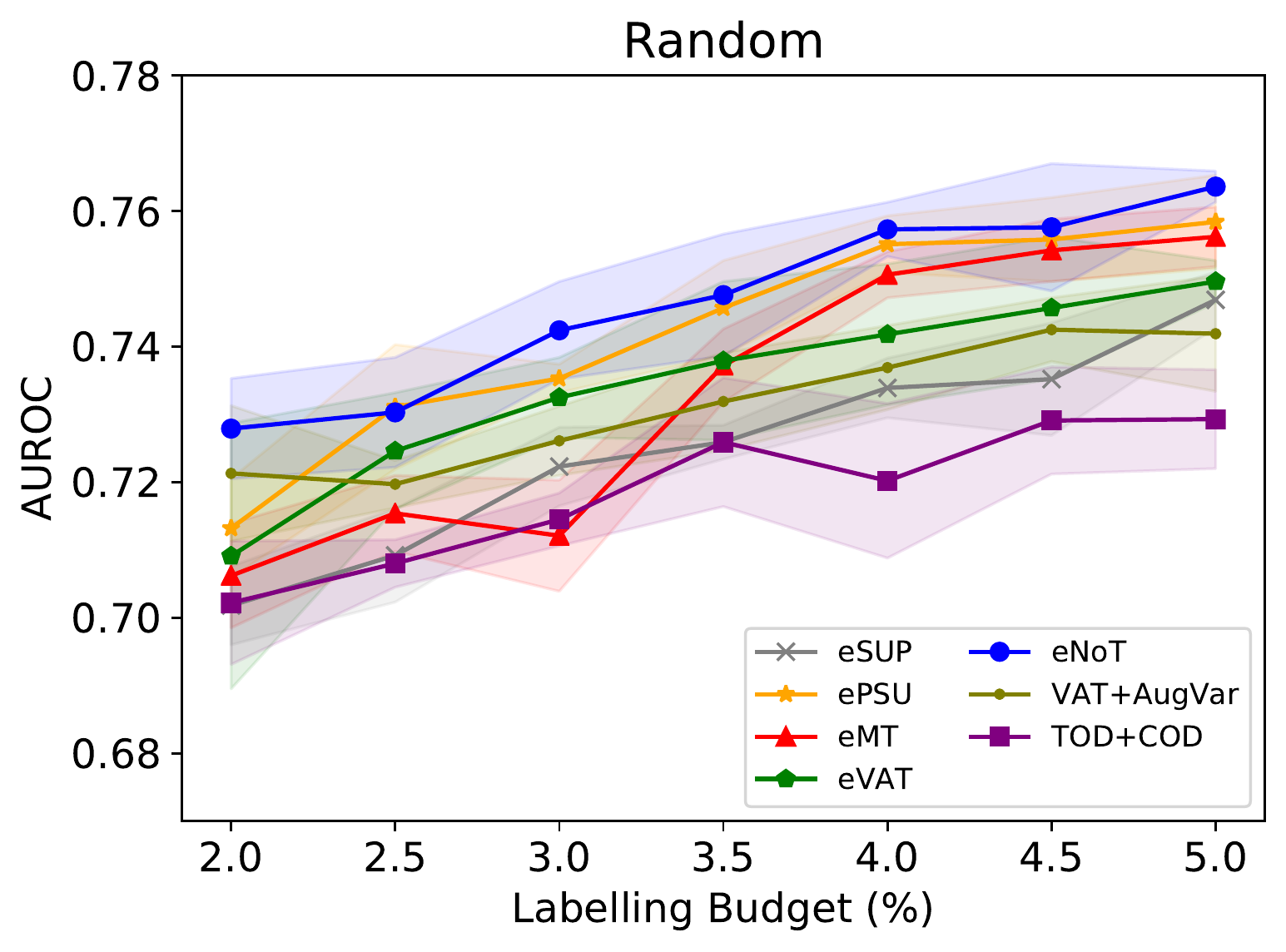}
\includegraphics[width=0.49\textwidth]{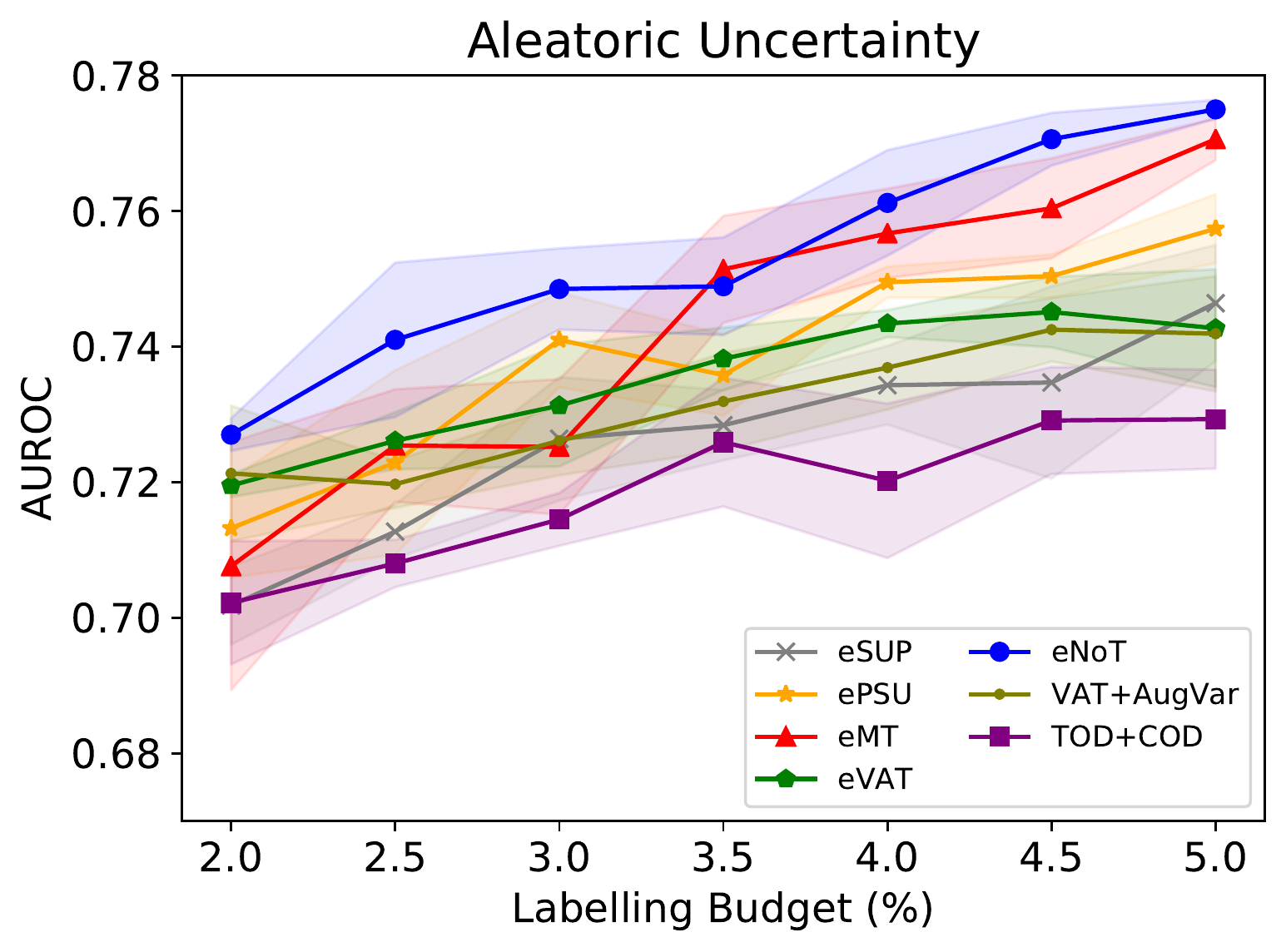}
\centering
\caption{Average test AUROC (mean $\pm$ std across 5 runs) on the NIH-14 Chest X-Ray dataset with different labelling budgets in the low-range labelling regime for random sampling (\textit{left}) and aleatoric uncertainty (\textit{right}). The proposed evidential counterparts of Pseudo-labelling (ePSU), Virtual Adversarial Training (eVAT), Mean Teacher (eMT) and NoTeacher (eNoT) are compared against the semi-supervised active learning baselines, VAT+AugVar and TOD+COD.}
\label{fig:nih_avg_results}
\end{figure}

\noindent \textbf{Average AUC vs. Labelling Budget:} Figure~\ref{fig:nih_avg_results} shows the average test AUROC obtained as a function of the labelling budget in the low-range labelling regime. In most cases, the AU-based active learning methods outperform their random sampling based counterparts. In particular, the performance gains under the best CSEAL method are higher in early annotation rounds, wherein the evidential active supervised method (eSUP+AU) could be biased on account of starting with very small randomly sampled labelled sets. This suggests that the integration of semi-supervised and active learning under CSEAL offers increased robustness to the cold-start problem~\cite{konyushkova2017learning}. Amongst the 4 semi-supervised evidential active learning methods, eNoT outperforms the others, possibly on account of its better consistency enforcement mechanism. Finally, we find that the best performing CSEAL method eNoT+AU outperforms the competing semi-supervised active learning baselines, TOD+COD and VAT+AugVar, by up to 5.4\% in AUROC and up to 2.6\% in AUPRC on average at the final labelling budget of 5\%. 

\begin{figure}[h!]
\includegraphics[width=0.9\textwidth, trim={0.4cm 0.4cm 0.2cm 0cm}, clip]{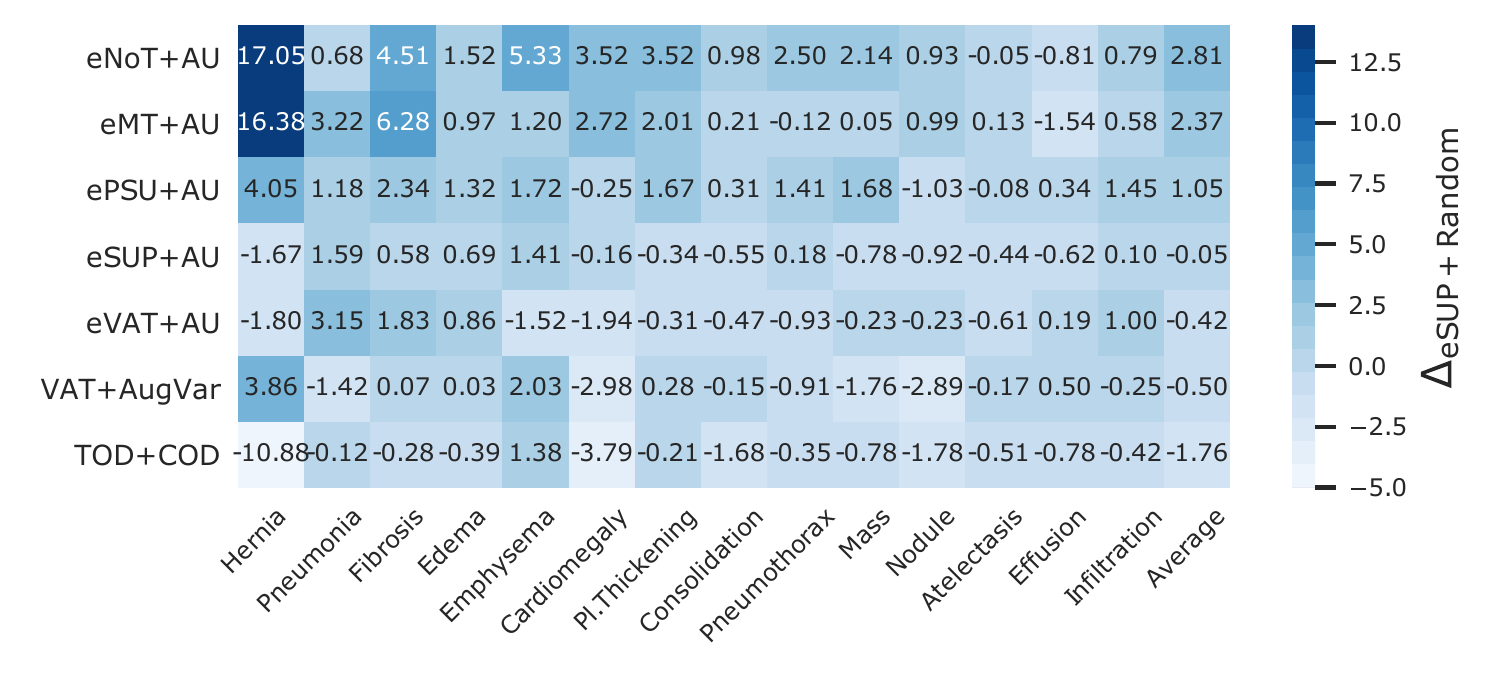} 
\centering
\caption{Class-wise breakdown of the AUROC gains (in \%) against the eSUP+Random baseline at the 5\% labelling budget, i.e., the end of the active learning process for the low-range labelling regime. All numbers in the heatmap are averaged over 5 runs. The proposed evidential counterparts of Pseudo-labelling (ePSU), Virtual Adversarial Training (eVAT), Mean Teacher (eMT) and NoTeacher (eNoT) with aleatoric uncertainty sampling (AU) are compared against the semi-supervised active learning baselines, VAT+AugVar and TOD+COD.}
\label{fig:nih_cw_results}
\end{figure}

\noindent \textbf{Analysis of Class-wise Performance Gains:}
At the end of the active learning process in the low-range labelling regime, i.e. at a labelling budget of 5\%, we compute the AUROC gain of all methods evaluated over eSUP+Random and plot the class-wise breakdowns of the AUROC gain in Figure~\ref{fig:nih_cw_results}. The classes are ordered by their prevalence in the test set. We observe that eMT+AU and eNoT+AU are comparable to or better than the competing baselines VAT+AugVar and TOD+COD. In particular, they have substantially larger gains over eSUP+Random for rarer classes. For example, eNoT+AU gains 17.05\% for Hernia and 3.52\% for Pleural Thickening which have a prevalence of $<0.2\%$ and $<3.3\%$ respectively. We posit that the enhanced performance of these CSEAL methods could be attributed to their consistency enforcement mechanism. Also, CSEAL estimates the aleatoric uncertainty as the informativeness criterion for active learning more effectively, especially for classes with few labelled samples.

%%%%%%%%%%%%%%%%%%%%%%%%%%%%%%%%%%%%%%%%%%%%%%%%%%%%%%%%

\section{Conclusions}
We introduced CSEAL, a novel end-to-end semi-supervised active learning framework for multi-label radiology image classification. We applied our framework to propose evidential analogues of 4 leading consistency-based semi-supervised learning methods, namely: evidential Pseudo-labelling (ePSU), evidential Virtual Adversarial Training (eVAT), evidential Mean Teacher (eMT) and evidential NoTeacher (eNoT). During active learning within a realistic annotation process, we demonstrate that CSEAL methods improve upon two leading semi-supervised active learning baselines. Amongst the CSEAL methods, we find that eNoT+AU provides the best performance across classes. Further, a class-wise breakdown shows that our best performing methods can gain up to 17\% in AUROC over the evidential supervised learning approaches for rarer classes (with $<5\%$ prevalence). Although our approach has currently been demonstrated on a single albeit challenging task of multi-label Chest X-Ray image classification with a specific convolutional backbone, it is amenable for future extensions to additional radiology modalities, such as CT and MRI, and networks. Further, future work could focus on investigating the different loss components empirically and theoretically to provide insight into the specific factors contributing to the effectiveness of CSEAL.

%%%%%%%%%%%%%%%%%%%%%%%%%%%%%%%%%%%%%%%%%%%%%%%%%%%%%%%%
\bigskip \small{\noindent\textbf{Acknowledgements:} Research efforts were supported by the Singapore International Graduate Award (SINGA Award to Shafa Balaram), Agency for Science, Technology and Research (A*STAR), Singapore; the A*STAR AME Programmatic Funds (Grant No. A20H6b0151); as well as funding and infrastructure for deep learning and medical imaging R\&D from the Institute for Infocomm Research, Science and Engineering Research Council, A*STAR}. We also thank Mike Shou from the National University of Singapore for his support and insightful discussions.

\bibliographystyle{splncs04}
\bibliography{references}

\begin{thebibliography}{10}
\providecommand{\url}[1]{\texttt{#1}}
\providecommand{\urlprefix}{URL }
\providecommand{\doi}[1]{https://doi.org/#1}

\bibitem{ash2019deep}
Ash, J.T., Zhang, C., Krishnamurthy, A., Langford, J., Agarwal, A.: Deep batch
  active learning by diverse, uncertain gradient lower bounds. In:
  International Conference on Learning Representations (2019)

\bibitem{borsos2021semi}
Borsos, Z., Tagliasacchi, M., Krause, A.: Semi-supervised batch active learning
  via bilevel optimization. In: ICASSP 2021-2021 IEEE International Conference
  on Acoustics, Speech and Signal Processing (ICASSP). pp. 3495--3499. IEEE
  (2021)

\bibitem{boushehri2021systematic}
Boushehri, S.S., Qasim, A., Waibel, D., Schmich, F., Marr, C.: Systematic
  comparison of incomplete-supervision approaches for biomedical imaging
  classification  (2021)

\bibitem{dempster1968generalization}
Dempster, A.P.: A generalization of bayesian inference. Journal of the Royal
  Statistical Society: Series B (Methodological)  \textbf{30}(2),  205--232
  (1968)

\bibitem{flanders2020construction}
Flanders, A.E., Prevedello, L.M., Shih, G., Halabi, S.S., Kalpathy-Cramer, J.,
  Ball, R., Mongan, J.T., Stein, A., Kitamura, F.C., Lungren, M.P., et~al.:
  Construction of a machine learning dataset through collaboration: the {RSNA}
  2019 brain {CT} hemorrhage challenge. Radiology: Artificial Intelligence
  \textbf{2}(3),  e190211 (2020)

\bibitem{gal2016uncertainty}
Gal, Y.: Uncertainty in deep learning. University of Cambridge. Ph.D. thesis,
  PhD Thesis. 2016.[2020-12-07]. http://mlg. eng. cam. ac. uk/yarin/thesis~…
  (2016)

\bibitem{gao2020consistency}
Gao, M., Zhang, Z., Yu, G., Ar{\i}k, S.{\"O}., Davis, L.S., Pfister, T.:
  Consistency-based semi-supervised active learning: Towards minimizing
  labeling cost. In: European Conference on Computer Vision. pp. 510--526.
  Springer (2020)

\bibitem{ghesu2021quantifying}
Ghesu, F.C., Georgescu, B., Mansoor, A., Yoo, Y., Gibson, E., Vishwanath, R.,
  Balachandran, A., Balter, J.M., Cao, Y., Singh, R., et~al.: Quantifying and
  leveraging predictive uncertainty for medical image assessment. Medical Image
  Analysis  \textbf{68},  101855 (2021)

\bibitem{guo2021semi}
Guo, J., Shi, H., Kang, Y., Kuang, K., Tang, S., Jiang, Z., Sun, C., Wu, F.,
  Zhuang, Y.: Semi-supervised active learning for semi-supervised models:
  exploit adversarial examples with graph-based virtual labels. In: Proceedings
  of the IEEE/CVF International Conference on Computer Vision. pp. 2896--2905
  (2021)

\bibitem{huang2021semi}
Huang, S., Wang, T., Xiong, H., Huan, J., Dou, D.: Semi-supervised active
  learning with temporal output discrepancy. In: Proceedings of the IEEE/CVF
  International Conference on Computer Vision. pp. 3447--3456 (2021)

\bibitem{jsang2016subjective}
Jsang, A.: Subjective logic: A formalism for reasoning under uncertainty.
  Springer Verlag  (2016)

\bibitem{konyushkova2017learning}
Konyushkova, K., Sznitman, R., Fua, P.: Learning active learning from data.
  Advances in neural information processing systems  \textbf{30} (2017)

\bibitem{lee2013pseudo}
Lee, D.H.: Pseudo-label: The simple and efficient semi-supervised learning
  method for deep neural networks. In: ICML Workshop on challenges in
  representation learning. vol.~3 (2013)

\bibitem{mckinney2020international}
McKinney, S.M., Sieniek, M., Godbole, V., Godwin, J., Antropova, N., Ashrafian,
  H., Back, T., Chesus, M., Corrado, G.S., Darzi, A., et~al.: International
  evaluation of an {AI} system for breast cancer screening. Nature
  \textbf{577}(7788),  89--94 (2020)

\bibitem{miyato2018virtual}
Miyato, T., Maeda, S.i., Koyama, M., Ishii, S.: Virtual adversarial training: a
  regularization method for supervised and semi-supervised learning. IEEE
  transactions on pattern analysis and machine intelligence  \textbf{41}(8),
  1979--1993 (2018)

\bibitem{paszke2017automatic}
Paszke, A., Gross, S., Chintala, S., Chanan, G., Yang, E., DeVito, Z., Lin, Z.,
  Desmaison, A., Antiga, L., Lerer, A.: Automatic differentiation in
  {P}y{T}orch  (2017)

\bibitem{rajpurkar2018deep}
Rajpurkar, P., Irvin, J., Ball, R.L., Zhu, K., Yang, B., Mehta, H., Duan, T.,
  Ding, D., Bagul, A., Langlotz, C.P., et~al.: Deep learning for chest
  radiograph diagnosis: A retrospective comparison of the {C}he{XN}e{X}t
  algorithm to practicing radiologists. Public Library of Science Medicine
  \textbf{15}(11),  e1002686 (2018)

\bibitem{sensoy2018evidential}
Sensoy, M., Kaplan, L., Kandemir, M.: Evidential deep learning to quantify
  classification uncertainty. Advances in Neural Information Processing Systems
   \textbf{31} (2018)

\bibitem{song2019combining}
Song, S., Berthelot, D., Rostamizadeh, A.: Combining mixmatch and active
  learning for better accuracy with fewer labels. arXiv preprint
  arXiv:1912.00594  (2019)

\bibitem{tarvainen2017mean}
Tarvainen, A., Valpola, H.: Mean teachers are better role models:
  Weight-averaged consistency targets improve semi-supervised deep learning
  results. In: Advances in Neural Information Processing Systems. pp.
  1195--1204 (2017)

\bibitem{unnikrishnan2021semi}
Unnikrishnan, B., Nguyen, C., Balaram, S., Li, C., Foo, C.S., Krishnaswamy, P.:
  Semi-supervised classification of radiology images with {N}o{T}eacher: A
  teacher that is not mean. Medical Image Analysis  \textbf{73},  102148 (2021)

\bibitem{unnikrishnan2020semi}
Unnikrishnan, B., Nguyen, C.M., Balaram, S., Foo, C.S., Krishnaswamy, P.:
  Semi-supervised classification of diagnostic radiographs with {N}o{T}eacher:
  a teacher that is not mean. In: International Conference on Medical Image
  Computing and Computer-Assisted Intervention. pp. 624--634. Springer (2020)

\bibitem{wang2016cost}
Wang, K., Zhang, D., Li, Y., Zhang, R., Lin, L.: Cost-effective active learning
  for deep image classification. IEEE Transactions on Circuits and Systems for
  Video Technology  \textbf{27}(12),  2591--2600 (2016)

\bibitem{wang2017hospital}
Wang, X., Peng, Y., Lu, L., Lu, Z., Bagheri, M., Summers, R.: Hospital-scale
  chest {X}-ray database and benchmarks on weakly-supervised classification and
  localization of common thorax diseases. In: IEEE Conference on Computer
  Vision and Pattern Recognition. pp. 3462--3471 (2017)

\bibitem{yang2020deep}
Yang, Z., Wu, W., Zhang, J., Zhao, Y., Gu, L.: Deep co-training active learning
  for mammographic images classification. In: 2020 Chinese Automation Congress
  (CAC). pp. 1059--1062. IEEE (2020)

\bibitem{zech284reproduce}
Zech, J.: reproduce-chexnet. GitHub repository (2018),
  \url{https://github.com/jrzech/reproduce-chexnet}

\bibitem{zhao2020uncertainty}
Zhao, X., Chen, F., Hu, S., Cho, J.H.: Uncertainty aware semi-supervised
  learning on graph data. Advances in Neural Information Processing Systems
  \textbf{33},  12827--12836 (2020)

\bibitem{zotova2019comparison}
Zotova, D., Lisowska, A., Anderson, O., Dilys, V., O’Neil, A.: Comparison of
  active learning strategies applied to lung nodule segmentation in {CT} scans.
  In: Large-Scale Annotation of Biomedical Data and Expert Label Synthesis and
  Hardware Aware Learning for Medical Imaging and Computer Assisted
  Intervention, pp. 3--12. Springer (2019)

\end{thebibliography}
%

%%%%%%%%%%%%%%%%%%%%%%%%%%%%%%%%%%%%%%%%%%%%%%%%%%%%%%%%
\section*{Supplementary Material}

% change caption name to include 'S' and reset counters
\renewcommand*{\thesection}{S\arabic{section}}
\renewcommand*{\thesubsection}{S\arabic{subsection}}
\renewcommand*{\thesubsubsection}{S\arabic{subsection}.\arabic{subsubsection}}
\renewcommand{\thetable}{S\arabic{table}}  
\setcounter{table}{0}
\renewcommand{\thefigure}{S\arabic{figure}}
\setcounter{figure}{0}

\subsection{CSEAL Losses \& Implementation Details}

\begin{fleqn}
\label{eq:esup_loss}
\begin{align*}
    \mathfrak{L}_{\eSUP} &= \mathfrak{L}_{err}(\vy, \hat{\mathbf{p}}^\mathrm{L}) + \mathfrak{L}_{var}(\hat{\mathbf{p}}^\mathrm{L}, \valpha^\mathrm{L}) + \lambda_t\ \mathfrak{L}_{reg}(\valpha^\mathrm{L}, \vy) \\
    &\approx (y^+ - \hppos)^2 + (y^- - \hpneg)^2 + \frac{\hppos(1-\hppos) + \hpneg(1-\hpneg)}{E+1} + \lambda_t\ \left[ \log \frac{\mathrm{\Gamma}(\tilde{\alpha} + \tilde{\beta})}{\mathrm{\Gamma}(\tilde{\alpha})\mathrm{\Gamma}(\tilde{\beta})} + \sum_{\gamma \in \{\tilde{\alpha}, \tilde{\beta}\}} \left(\gamma - 1\right) \left(\psi(\gamma) - \psi(\tilde{\alpha} + \tilde{\beta}) \right) \right]
\end{align*}
\end{fleqn} 
    
\begin{fleqn}
\label{eq:epsu_loss}
\begin{align*}
    \mathfrak{L}_{\ePSU} = \mathfrak{L}_{\eSUP}(\vxl, \vy) + \mathfrak{L}_{\eSUP}(\vxu, \vy_{psu}) 
\end{align*}
\end{fleqn} 

\begin{fleqn}
\label{eq:evat_loss}
\begin{align*}
    \mathfrak{L}_{\mathrm{eVAT}} &= \mathfrak{L}_{\eSUP}(\vxl, \vy) + \lambda_{cons}\ \left[\mathfrak{L}_{err}(\vhp^\mathrm{U}, \vhp^\mathrm{U}_{adv}) + \mathfrak{L}_{var}(\vhp^\mathrm{U}, \valpha^\mathrm{U}) + \mathfrak{L}_{var}(\vhp^\mathrm{U}_{adv}, \valpha^\mathrm{U}_{adv}) \right] 
\end{align*}
\end{fleqn}

\begin{fleqn}
\label{eq:emt_loss}
\begin{align*}
    \mathfrak{L}_{\mathrm{eMT}} &=
    \mathfrak{L}_{\eSUP}(\vxl_S, \vy) + \lambda_{cons}\ \left[\mathfrak{L}_{err}(\vhp_S, \vhp_T) + \mathfrak{L}_{var}(\vhp_S, \valpha_S) + \mathfrak{L}_{var}(\vhp_T, \valpha_T) \right]
\end{align*}
\end{fleqn}

\begin{fleqn}
\label{eq:enot_loss}
\begin{align*}
    \mathfrak{L}_{\eNoT} 
    &= \lambda_{sup,1} \ \mathfrak{L}_{\eSUP}(\vxl_1, \vy) + \lambda_{sup,2}  \  \mathfrak{L}_{\eSUP}(\vxl_2, \vy) + \lambda_{cons}^\mathrm{L}  \ \left[\mathfrak{L}_{err}(\vhponel, \vhptwol) + \mathfrak{L}_{var}(\vhponel, \valphaonel) +  \mathfrak{L}_{var}(\vhptwol, \valphatwol) \right] + \\
     &\ \ \ \lambda_{cons}^\mathrm{U} \left[\mathfrak{L}_{err}(\vhptwou, \vhptwou) + \mathfrak{L}_{var}(\vhptwou, \valphatwou) +  \mathfrak{L}_{var}(\vhptwou, \valphatwou) \right] 
\end{align*}
\end{fleqn} 
where the subscripts $psu$, $adv$, $S$ and $T$ refer to the pseudo-labelled samples, the adversarial examples of the unlabelled samples, and the student and teacher models respectively. 

\vspace{2ex}
\begin{table}[h!]
\caption{Hyperparameter values for the CSEAL methods and baselines, for each of the low-range and mid-range labelling regimes. The hyperparameters $\alpha$, $N$ and $\lambda$ correspond to the EMA decay rate, number of augmented samples and the trade-off weight. The dropout rate $p$ is set to $0.50$ unless otherwise specified. The early stopping and reduce learning rate patience are 15 and 5 epochs respectively. }
\label{tab1:hyperparameters}
\resizebox{\textwidth}{!}{
\begin{tabular}{|l|l|c|c|c|c|c|c|c|c|c|c|c|c|c|c|c|c|}
\hline
\multicolumn{2}{|l|}{} & \textbf{eSUP} & \multicolumn{2}{c|}{\textbf{eVAT}} & \multicolumn{3}{c|}{\textbf{eMT}} & \multicolumn{5}{c|}{\textbf{eNoT}} & \multicolumn{2}{c|}{\textbf{VAT+ }} & \multicolumn{3}{c|}{\textbf{TOD+}} \\ 
 \multicolumn{2}{|l|}{} & & \multicolumn{2}{c|}{} & \multicolumn{3}{c|}{}  & \multicolumn{5}{c|}{} & \multicolumn{2}{c|}{\textbf{AugVar}} & \multicolumn{3}{c|}{\textbf{COD}}\\ 
\cline{3-18} 
\multicolumn{2}{|l|}{} & $\lambda_{sup}$ & $\lambda_{cons}$ & $p$ &  $\lambda_{cons}$ & $\alpha$ & $p$ & $\lambda_{sup,1}$ & $\lambda_{sup,2}$ & $\lambda_{cons}^\mathrm{L}$ & $\lambda_{cons}^\mathrm{U}$ & $p$ & $N$ & $p$ & $\alpha$ & $\lambda$ & $p$\\
\hline 
\multirow{2}{*}{Low} & Random & \multirow{2}{*}{1} & \multirow{2}{*}{1} & 0.20 & \multirow{2}{*}{196} & \multirow{2}{*}{0.91} & 0.30 & \multirow{2}{*}{0.67} & \multirow{2}{*}{0.67} & \multirow{2}{*}{0.67} & \multirow{2}{*}{1} & 0.20 & \multirow{2}{*}{10} & \multirow{2}{*}{0} & \multirow{2}{*}{0.999} & \multirow{2}{*}{0.05} & \multirow{2}{*}{0} \\ 
 & AU & & & 0.25 & & & 0.20 & & & & & 0.25 & & & & & \\ 
 \cline{1-18}
\multirow{2}{*}{Mid} & Random & \multirow{2}{*}{1} & \multirow{2}{*}{1} & \multirow{2}{*}{0.20} & \multirow{2}{*}{196} & \multirow{2}{*}{0.91} & \multirow{2}{*}{0.20} & \multirow{2}{*}{0.67} & \multirow{2}{*}{0.67} & \multirow{2}{*}{0.67} & \multirow{2}{*}{1} & \multirow{2}{*}{0.20} & \multirow{2}{*}{10} & \multirow{2}{*}{0} & \multirow{2}{*}{0.999} & \multirow{2}{*}{0.05} & \multirow{2}{*}{0} \\
 & AU & & & & & & & & & & & & & & & & \\
\hline
\end{tabular}}
\end{table} 
 
\noindent \textbf{Evidential vs. Non-Evidential Semi-supervised Learning Approaches:} 
We compare the gains in average test AUROC of the proposed evidential semi-supervised learning approaches over their non-evidential counterparts. For PSU, VAT, MT, and NoT, the evidential semi-supervised method gains, on average $[-0.03, -1.08, +1.81, +0.45]\%$ in AUROC at the the end of the low-range (budget $5\%$), respectively; and on average $[+0.45, -0.17, -0.07, -0.40]\%$ in AUROC at the end of the mid-range (budget $10\%$) labelling regimes, respectively.

\newpage 

\subsection{Mid-Range Labelling Regime Results}

\begin{figure}[h!]
\includegraphics[width=0.495\textwidth]{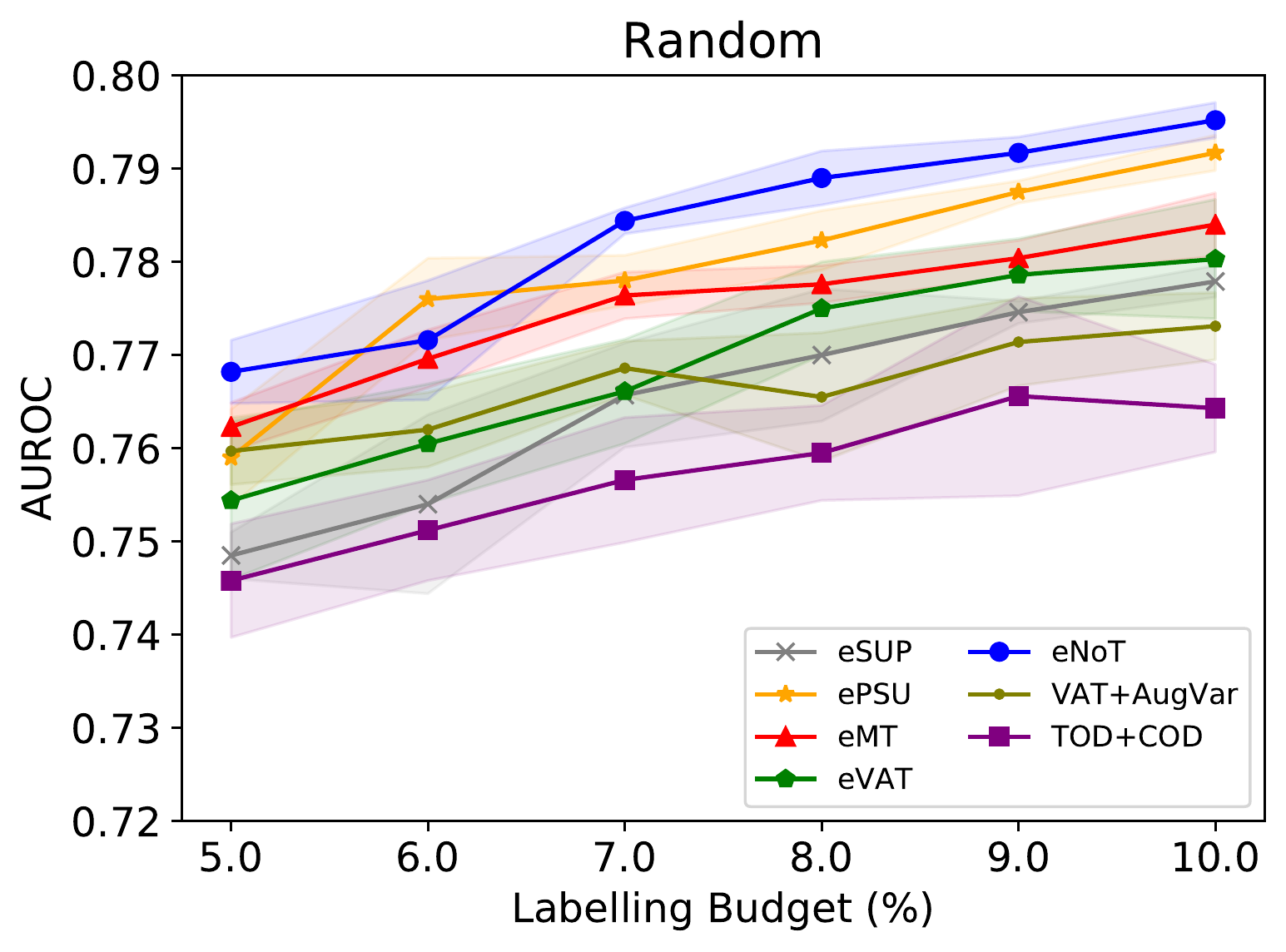}
\includegraphics[width=0.495\textwidth]{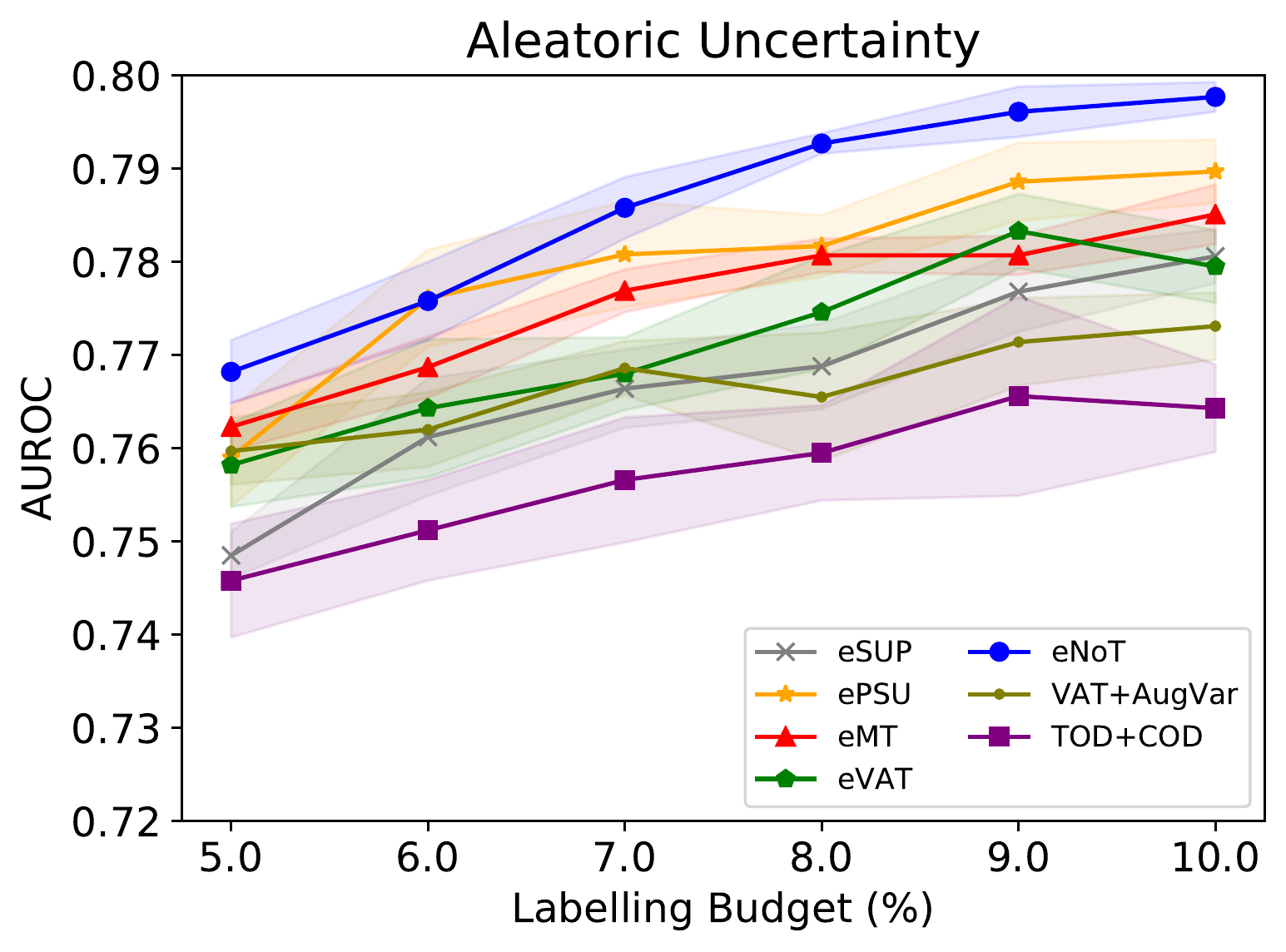}
\centering
\caption{Average test AUROC (mean $\pm$ std across 5 runs) on the NIH-14 Chest X-Ray dataset in the mid-range labelling regime for random sampling (\textit{left}) and aleatoric uncertainty, AU (\textit{right}). The proposed ePSU, eVAT, eMT and eNoT methods are compared against the semi-supervised active learning baselines, VAT+AugVar and TOD+COD. Similar to the low-range, CSEAL methods outperform the baselines substantially and eNoT remains the best-performing method. At the labelling budget of 10\%, eNoT+AU is up to 1.9\% higher in AUPRC than the baselines on average. That said, the impact of the sampling mechanism on performance in the mid-range is lower, possibly due to the higher abundance of labelled images.}
\label{fig:nih_avg_results_mid}
\end{figure}

\begin{figure}[h!]
\includegraphics[width=0.90\textwidth, trim={0.4cm 0.4cm 0.2cm 0cm}, clip]{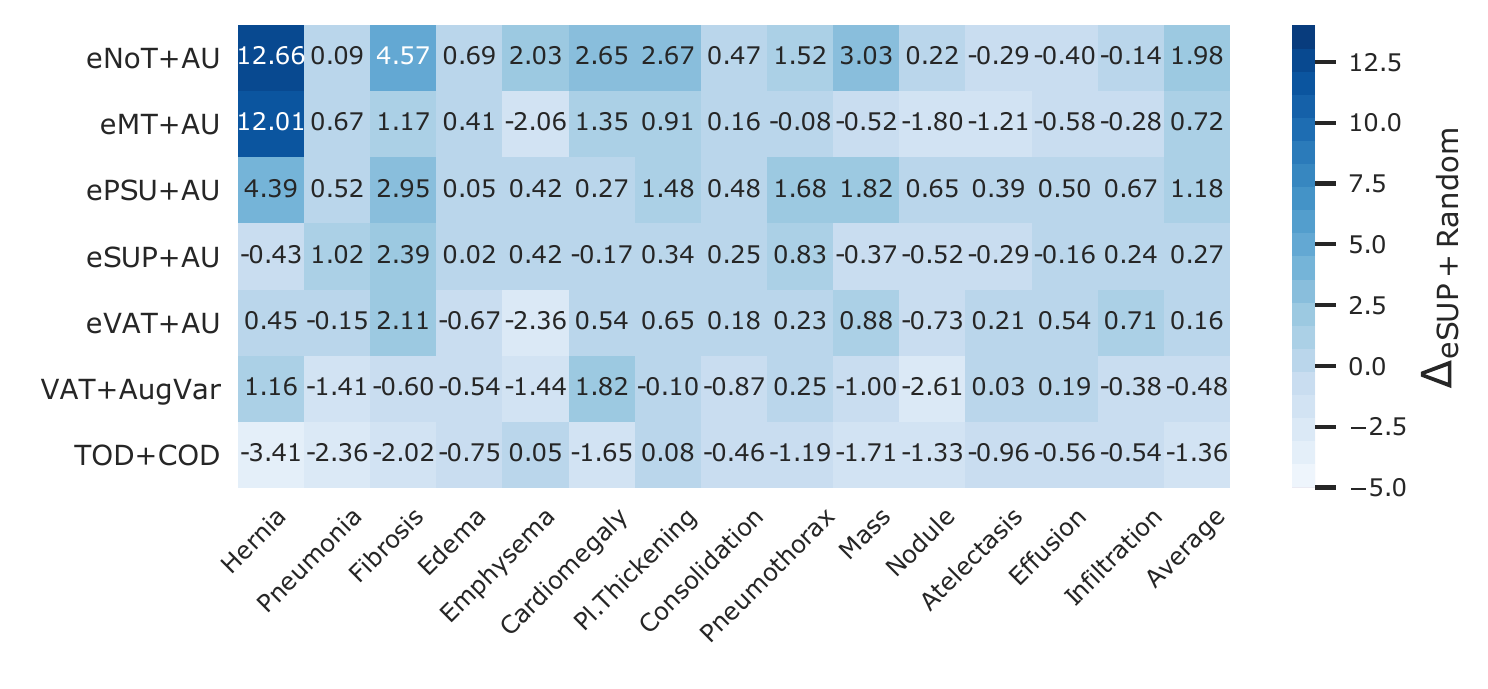} 
\centering
\caption{Class-wise breakdown of the AUROC gains (in \%) against the eSUP+Random baseline at the 10\% labelling budget, i.e., at the end of the active learning process for the mid-range labelling regime. All numbers in the heatmap are averages computed over 5 runs. The classes are ordered by their prevalence in the test set. The proposed ePSU, eVAT, eMT and eNoT with AU methods are compared against the semi-supervised active learning baselines, VAT+AugVar and TOD+COD. As observed for the low-range labelling regime, the CSEAL methods gain more over the baselines, especially for the rarer classes.}
\label{fig:nih_cw_results_mid}
\end{figure}

\end{document}